%% file: main.tex
\pdfoutput=1

\documentclass[11pt]{article}

\usepackage[table,xcdraw]{xcolor}
\usepackage[normalem]{ulem}
\useunder{\uline}{\ul}{}

\usepackage[]{acl}

\usepackage{times}
\usepackage{latexsym}

\usepackage[T1]{fontenc}

\usepackage[utf8]{inputenc}

\usepackage{microtype}

\usepackage{booktabs}

\usepackage{amsfonts}
\usepackage{bm}
\usepackage{subcaption}
\usepackage{mathtools}
\usepackage{hyperref}
\usepackage{balance}
\usepackage{longtable}
\usepackage[inline]{enumitem}

\usepackage{amssymb}
\DeclareMathOperator*{\argmax}{arg\,max}

\definecolor{cyan}{HTML}{567FC4}
\definecolor{purple}{HTML}{C95244}
\newcommand{\hashtag}[1]{\textcolor{cyan}{${#1}$}}
\newcommand{\user}[1]{\textcolor{purple}{${#1}$}}
\newcommand{\ntulm}[0]{\texttt{NTULM}}

\title{NTULM: Enriching Social Media Text Representations\\with Non-Textual Units}

\makeatletter
\def\thanks#1{\protected@xdef\@thanks{\@thanks
        \protect\footnotetext{#1}}}
\makeatother

\author{
Jinning Li$^{1,2\S}$ \thanks{~~ $^{\S}$Equal Contribution. Corresponding Author: \texttt{smishra@twitter.com}},
Shubhanshu Mishra$^{1\S}$,
Ahmed El-Kishky$^{1}$,
Sneha Mehta$^{1}$,
Vivek Kulkarni$^{1}$
\\
$^{1}$ Twitter, Inc. \\
$^{2}$ University of Illinois at Urbana-Champaign \\
\texttt{jinning4@illinois.edu} \\
\texttt{\{smishra, aelkishky, snehamehta, vkulkarni\}@twitter.com} \\
}

\begin{document}
\maketitle
\input{00_abstract}
\input{01_intro}
\input{02_formulation}

\input{03_model}

\input{04_experiment}
\input{05_relatedwork}
\input{06_conclusion}


\bibliography{main}

\clearpage
\appendix
\input{07_appendix}
\end{document}

%% file: 00_abstract.tex
\begin{abstract}
On social media, additional context is often present in the form of annotations and meta-data such as the post's author, mentions, Hashtags, and hyperlinks. We refer to these annotations as Non-Textual Units (NTUs). We posit that NTUs provide social context beyond their textual semantics and leveraging these units can enrich social media text representations. In this work we construct an NTU-centric social heterogeneous network to co-embed NTUs. We then principally integrate these NTU embeddings into a large pretrained language model by fine-tuning with these additional units. This adds context to noisy short-text social media. Experiments show that utilizing NTU-augmented text representations significantly outperforms existing text-only baselines by 2-5\% relative points on many downstream tasks highlighting the importance of context to social media NLP. We also highlight that including NTU context into the initial layers of language model alongside text is better than using it after the text embedding is generated. Our work leads to the generation of holistic general purpose social media content embedding. 
\end{abstract}

%% file: 01_intro.tex
\section{Introduction}

\begin{figure*}
    \centering
    \includegraphics[width=\linewidth]{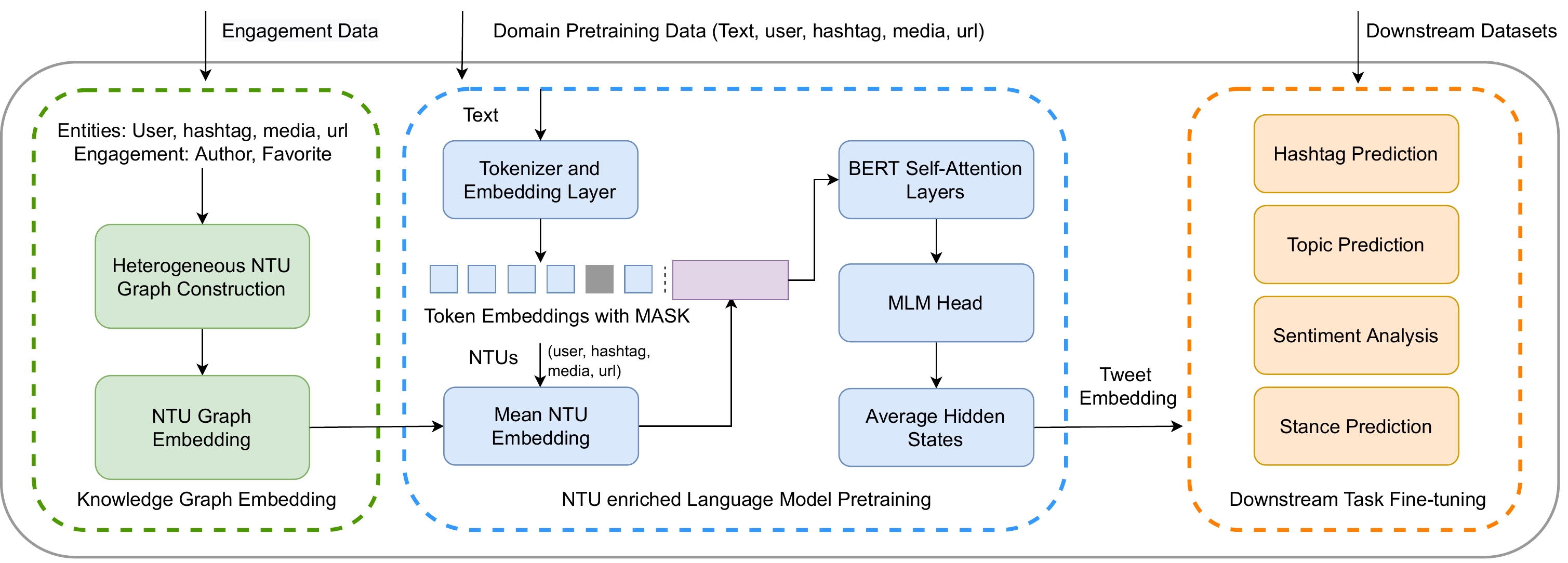}
    \caption{The framework of \ntulm{} model. In the Knowledge Graph Embedding module, we use the engagement data to build the heterogeneous graph and train large-scale NTU embeddings. In the NTU enriched LM pre-training, we incorporate the mean NTU embedding at the end of the sequence. We compute the tweet embedding as the average of the last hidden states and use it for multiple downstream tasks.}
    \label{fig:kelm}
\end{figure*}

Understanding the social context is crucial to the semantic understanding of a social media post~\cite{nguyen2016computational,kulkarni-etal-2021-lmsoc-approach,Mishra2018UserTextCorr,hovy-2015-demographic}. This is especially true for short-text social media such as Twitter where the textual content available for semantic understanding is inherently limited. As such, pretrained language models that ignore non-textual context can demonstrate sub-optimal performance when utilized for social-media NLP.

Fortunately, on social media, there are many available non-textual units (NTUs), which provide social contexts for any written text. For example, the author of a post provides a social prior as to the content written by that author. Additionally, the author may annotate their post with meta-data such as Hashtags, user mentions, or URLs and other media. These units can frame the content of a post by providing social context, a stance, or additional supporting material.

Previous research has investigated augmenting pretrained language model representations with additional signals. These include enrichments by incorporating image features~\cite{sun2020riva}, better-segmented Hashtags~\cite{maddela2019multi}, URL understanding~\cite{yasunaga2022linkbert}, or temporal-spatial contexts~\cite{kulkarni-etal-2021-lmsoc-approach}. 

However, these existing works are type-specific and require a specialized technique to integrate just one type of non-textual signal (e.g., requiring an image encoder to extract image features). We claim that this added complexity makes it difficult to incorporate different non-textual signals and effectively train a joint model. 

In this paper, our NTU enriched Language Model (\ntulm{}) can easily, without loss of generality, train and integrate graph embeddings~\cite{el2022graph} for \textit{multiple} types of NTUs. \ntulm{} can do this through the use of heterogeneous information network embeddings of NTUs. This allows us to not only co-embed multiple NTU types, but also incorporate a variety of interaction types as edges in our network (e.g., \textit{authoring} posts, \textit{favoriting} Hashtags, and \textit{co-mentioning} users). This general embedding framework is simple and does not require specialized feature encoders for different NTU types. After obtaining the NTU knowledge embeddings, \ntulm{} deeply integrates them with the language model at the token level and simply applies the default attention mechanism used in the BERT encoder. To ensure our alignment with \cite{kulkarni-etal-2021-lmsoc-approach} which allows only inclusion of a single context embedding to BERT, we take the average of NTU embeddings and attach the unified embedding at the end of token embedding sequence. The framework of \ntulm{} is shown in Figure~\ref{fig:kelm}.

To ensure high coverage of the NTU vocabulary across tweets, we construct a large-scale heterogeneous NTU graph ensuring high overlap with all tweets. With the scalability of our graph embeddings, we can rapidly embed NTUs ensuring high coverage across tweets.

We state and analyze the problem in Section~\ref{sec:formulation}, followed
by our proposed solution involving NTU embedding and BERT integration in Section~\ref{sec:model}. In Section~\ref{sec:exp} we evaluate our proposed solution compared to text-only baselines. We go over related works in Section~\ref{sec:relatedwork} and conclude in Section~\ref{sec:conclude}.

%% file: 02_formulation.tex
\section{Task Formulation}\label{sec:formulation}
In this section, we formulate the task of enriching pretrained language models with additional NTU embeddings.

\subsection{Non-Textual Units (NTUs)}
Social media posts are composed of textual content and non-textual units (NTUs) which provide additional context to the text. These include: the author of a post, any mentioned users, annotated topics via Hashtags, shared URLs, etc. While some of these units are encoded textually within a post, their meaning is not fully encapsulated by their textual semantics. Instead, this meaning can be better derived by understanding the social community that engages with the NTUs. Take for example the Hashtag \hashtag{nlproc} which is used by the Natural Language Processing community; this differs from \hashtag{nlp} which is used by the natural language processing community \textit{and} the Neuro-linguistic programming community. While both Hashtags contain the subword \texttt{nlp}, the real meaning is dependent on the social context they occur (e.g., from the author and social Hashtag embedding). This problem is more difficult with user mentions which convey no linguistic information in their textual form but can be more informative if mentions are considered by the social graph context of the user mentioned. We represent these NTUs using the heterogeneous social graph where each NTU is a node, and multi-typed edges represent their relation to other NTUs.

\subsection{Integrating NTUs in Language Models}
We extend the work introduced by LMSOC \cite{kulkarni-etal-2021-lmsoc-approach}, which demonstrates that the integration of temporal and geographical context in Tweet texts leads to better performance on cloze tasks. Similar to LMSOC, we take a base language model and integrate the NTU information in this model as additional context. Our goal is that each token in the text should not just be contextualized by other tokens in the text but also by the NTUs associated with the text. This approach is generic and we describe the exact choice of language model and NTU integration in detail later. 

We improve on LMSOC by:
\begin{enumerate*}[label=(\roman*)]
\item learning richer representations for NTUs using Heterogeneous Information Network embedding approaches~\cite{TwHIN}, 
\item usage of social engagement signals, 
\item utilizing multiple tweet contexts via multiple NTU embeddings, 
\item assessing the performance of these models on a wide variety of downstream Tweet classification tasks
\end{enumerate*}.

Finally, we propose a holistic and end-to-end pipeline for training models with NTUs.

%% file: 03_model.tex
\section{NTU enriched Language Model}
\label{sec:model}
The framework of \ntulm{} is shown in Figure~\ref{fig:kelm}. In this section, We first introduce how we learn high-quality NTU embeddings by embedding an NTU-centric heterogenous social graph. We then describe how we principally integrate these NTU embeddings in a standard BERT-style language model yielding Tweet embeddings that utilize both text and NTU information. 

We will use the Tweet in Table~\ref{table::exampletweet} as an example for the following sections.
\begin{table}[htb]
\begin{tabular}{|l|}
\hline
\textbf{Author}: \user{user1}\\
\textbf{Tweet}: Our paper was accepted at \user{@WNUT} \\
with \user{@user2} \user{@user3} \hashtag{\#nlproc} \hashtag{\#socialmedia}\\
\textbf{Favorited by}: \user{user4}, \user{user5}\\
\hline
\end{tabular}
\caption{Example tweet with engagement data of author, mentions, Hashtags, and favorites}\label{table::exampletweet}
\end{table}


\subsection{NTU Graph Construction and Embedding}
We seek to understand NTUs based on the social context in which they're engaged and construct a dense NTU representation such that similar NTUs are close in this dense embedding space.

\begin{figure}
    \centering
    \includegraphics[width=0.98\linewidth]{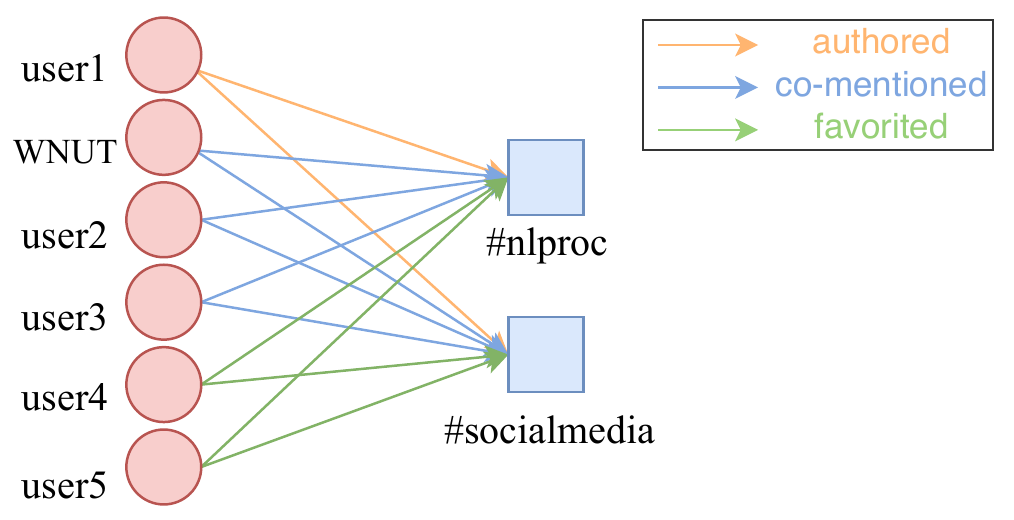}
    \caption{Graph construction with the example data in Table~\ref{table::exampletweet} for training \ntulm{} user-Hashtag embeddings.}
    \label{fig:kelm-twhin-graph}
\end{figure}

\paragraph{Constructing Heterogeneous Network:}
We start by constructing a large-scale heterogeneous graph $\mathcal{G}$ which models engagement between users and a set of NTU-observed Tweets (any language from 2018 till 2022). This heterogeneous graph consists of nodes and edges where multiple edges of different types can exist between a pair of nodes. For this work, we focus on users and Hashtags as NTUs, because they are the most accessible NTUs and are available or retrievable on most datasets. We construct the graph by taking a sample of Tweets, extracting the mention users, Hashtags, and the Tweet author. We also include a list of users who have favorited the Tweet. This leads to a graph where the nodes are either users or Hashtags. We include an edge between a user and a Hashtag if the user has either \textit{favorited} a Tweet with the Hashtag, \textit{authored} a Tweet with the Hashtag, or is \textit{co-mentioned} with a Hashtag. One example of constructed graph is provided in Figure~\ref{fig:kelm-twhin-graph}. Our choice of edges is based on the easy availability of the user Hashtag data via the Twitter API.

We construct this graph using data from January 1st, 2018 to July 1st, 2022. This leads to a graph with $60$M Hashtags, $255$M users, $5$B authorship edges, $3$B favorite edges, and $0.9$B co-mention edges. We then learn heterogeneous graph embeddings by following the approach outlined in TwHIN~\cite{TwHIN}. This gives us a set of embeddings for Users and Hashtags which exist in the same embedding space.

\paragraph{Heterogeneous Graph Embedding:} We learn embedding vectors by applying a similar scheme to TransE~\cite{bordes2013translating}. For a pair of nodes in the graph ($u_i$), ($v_j$), we denote their embeddings as $\mathbf{u_i}$ and $\mathbf{v_j}$ respectively. We denote an edge as a triplet $e=( u_i, r_k, v_j$) which consists of head and tail nodes ($u_i, v_j$) connected by a specific relation ($r_k$). We score these triplets  with a scoring function of the form $f(\mathbf{u_i}, \mathbf{r_k}, \mathbf{v_j})$ where  $\mathbf{r_k}$ is the relation embedding. Our training objective seeks to learn $\mathbf{e}$ parameters that maximize a log-likelihood constructed from the scoring function for $e \in \mathcal{G}$ and minimize for $e \notin \mathcal{G}$.

For simplicity, we apply a simple dot product comparison between node representations. For an edge $e=(u_i, r_k, v_j)$, this operation is defined by:
\begin{equation}
    \label{eq:scoring}
    f(e) = f(u_i, r_k, v_j) = \mathbf{u_i}^\intercal (\mathbf{v_j} + \mathbf{r_k})
\end{equation}

As seen in Equation~\ref{eq:scoring}, we co-embed all nodes in $ \mathcal{G}$  by translating the tail node by the specific relation vector and scoring their respective embedded representations via dot product. The task is then formulated as an edge (or link) prediction task. We consume the input graph $\mathcal{G}$ as a set of (node, relation, node) triplets of the form  $(u, r, v)$ which represent a link between nodes in the graph. The embedding training objective is to find node and relation representations that are useful for predicting which nodes are linked via that specific relation. While a softmax is a natural formulation to edge prediction, it is impractical due to the cost of computing the normalization over a large vocabulary of nodes. Following previous methods~\cite{mikolov_distributed_2013,goldberg2014word2vec}, negative sampling, a simplification of noise-contrastive estimation, can be used to learn the parameters. We therefore maximize the following negative sampling objective,
\begin{equation}
    \label{eq:objective}
    \argmax_{\mathbf{u}, \mathbf{r}, \mathbf{v}}\sum_{e \in \mathcal{G}} [  \log \sigma (f(e)) + \sum_{e' \in N(e)} \log \sigma (-f(e')) ]
\end{equation}
where: $N(u,r, v) = \{(u, r, v’): v’ \in \mathcal{I}\} \cup \{(u’,r, v): u' \in \mathcal{U}\}$.
Equation~\ref{eq:objective} represents the log-likelihood of predicting a binary ``real" or ``fake'' label for the set of edges in the network (real) along with a set of the ``fake'' negatively sampled edges. To maximize the objective, we learn $\mathbf{u}$, $\mathbf{r}$, and $\mathbf{v}$  parameters to differentiate positive edges from negative, unobserved edges. Negative edges are sampled by corrupting positive edges via replacing either the user or item in an edge pair with a negatively sampled user or item. As user-item interaction graphs are very sparse, randomly corrupting an edge in the graph is very likely to be a `negative' edge absent from the graph.

\subsection{Enriching Language Model with NTU Embeddings}
In this section, we explain how we integrate these embeddings into a language model. We build on the LMSOC framework~\cite{kulkarni-etal-2021-lmsoc-approach} to append NTU embeddings into the MLM model. However, unlike LMSOC, which has only one context embedding, we now may have multiple NTU embeddings for a given Tweet. Taking the example above, the NTUs for the Tweet are \user{user1}, \user{WNUT}, \user{user2}, \user{user3}, \user{user4}, \user{user5}, \hashtag{\#nlproc}, \hashtag{\#socialmedia}. For our experiments we only limit ourselves to author and hashtag NTUs, i.e. \user{user1}, \hashtag{\#nlproc}, \hashtag{\#socialmedia}. This leads to a choice we have to make for integrating these NTU embeddings into the Tweet text. For this work we simply utilize the average of the NTU embeddings to keep it aligned with the LMSOC framework. In future we also plan to experiment with the social contexts used in LMSOC. 

Our final NTU embedding for the Tweet becomes the average embeddings of all NTUs in the Tweet. Let us denote it by $e_{ntu}$. We concatenate this embedding to the BERT's subword embedding. For NTUs not present in our NTU embeddings we use the average embedding of all the NTUs in our embedding table as a placeholder embedding. We found using the average as opposed to a zero embedding was much more beneficial for downstream task improvements. Furthermore, for Tweets which have no NTUs we also use the average NTU embedding as a placeholder embedding. Given a Tweet text, we tokenize it using the language model tokenizer into a list of subwords, we extract the subword embeddings from the model to get a list of subword embeddings. Lets call these subword embeddings $[s_0, s_1, s_2, ..., s_n]$. 

Since, $e_{ntu}$ and $s_i$ are of different embedding sizes, we use a linear layer to project $e_{ntu}$ in the space of $s_i$ and get $s_{ntu}$. This linear layer is jointly trained during MLM fine-tuning. We do not add a position embedding to the NTU and we do not add a type embedding to the NTU. Finally, we get a new list of embeddings of the Tweet i.e. $S = [s_0, s_1, s_2, ..., s_n, s_{ntu}]$. We feed these embedding to the next layers of a pre-trained Language model. We call this model a NTU enriched Language Model (\ntulm{}). 

The above model is then trained using the Masked Language Modeling (MLM) task similar to BERT model \cite{devlin2018bert}. We use the same setup for training via the MLM objective by masking 15\% of the tokens. This translates to the model learning to predict the missing words by using the NTU's context. 

While our approach is agnostic to the choice of encoder, for all our experiments we train based on a \texttt{bert-base-uncased} model using the HuggingFace Transformers library.\footnote{\url{https://huggingface.co/bert-base-uncased}} We train the models till convergence for a max of 15 epochs on a dataset of 1M English Tweets (see appendix \ref{sec:training-details}).

\subsection{NTU-enriched Text Embeddings}
Once the above model is trained, we use it in downstream tasks. Traditionally pre-trained language models are utilized in downstream tasks is by fine-tuning. However, this setup is not suitable for low-cost inference where multiple downstream models utilize the Tweet features, as doing inference on the full large-scale language model is expensive and doing inference of multiple BERT models is prohibitive. Furthermore, having a single Tweet embedding for all downstream tasks trades off accuracy for computing cost and allows the usage of caching of these Tweet embeddings for multiple downstream tasks. Motivated by this we generate fixed-size Tweet embedding which integrates Text and NTU information. We compare it with a text-only Tweet embedding. We refer to these embeddings as $\texttt{embed}_{ntulm}$ embeddings. Given input embeddings $S = [s_0, s_1, s_2, ..., s_n, s_{ntu}]$ we pass it through a language model which outputs $Z = [z_0, z_1, z_2, ..., z_n, z_{ntu}]$ embeddings. Our \ntulm{} embedding is the average of $z_i$ embeddings, i.e. $\texttt{embed}_{ntulm} = \frac{\sum_i z_i}{size(Z)}$. We feed these embedding as input to the downstream models and add a set of MLP layers on top to get the final prediction for each downstream model discussed in the experiments below. Note, that during downstream task training the \ntulm{} is frozen and not updated.

%% file: 04_experiment.tex
\section{Experiments}
\label{sec:exp}

We conduct experiments on a variety of datasets and downstream tasks to highlight the utility of \ntulm{}. Additionally, we perform an ablation to measure the contribution of each type of NTU to the overall \ntulm{} performance.

\subsection{Downstream Datasets}

In order to evaluate the performance of our models, we select the following downstream datasets. We choose classification datasets for all our evaluations. The statistics about our datasets can be found in Table~\ref{tab:downstream-data-stats} in appendix.

\paragraph{Topic Prediction}
We use a dataset of Tweets annotated with Topics as described in \cite{kulkarni-etal-2022-ctm}. This dataset consists of each Tweet annotated with a set of topics. The task is defined as: given a topic-based Tweet, retrieve tweets from the same topic. The final evaluation is based on Mean Average Precision (MAP).

\paragraph{Hashtag Prediction}
We use a dataset of 1M Tweets with Hashtags. The Hashtag prediction task is formulated as removing a single Hashtag from the Tweet and trying to predict using the remaining information in a multi-class classification task. For this task, we consider the top 1000 Hashtags as prediction classes and remove them from the Tweets containing these Hashtags. We use an equal number of Tweets for each Hashtag for our training and test sets. We evaluate the performance of \ntulm{} and baselines using Recall @ 10.

\paragraph{SemEval Sentiment}
We use the SemEval Sentiment dataset from 2017. This dataset is released in the form of Tweet Ids and labels. We hydrate the Tweet ids using the public Twitter Academic API and fetch the author, Hashtags, and Tweet text from the API response. Because of the deletion of many Tweet ids we can not compare our results with previous baselines hence our only comparison is with the BERT-based baseline we consider. We use the macro F1 score as well. The SemEval dataset consists of three tasks. Task A consists of multi-class sentiment classification where given a Tweet we need to predict the label among positive, negative, and neutral. Task BD consists of topic-based sentiment prediction using only two classes positive, and negative. Task CE consists of Tweet quantification where we need to predict sentiment across a 5-point scale. For topic-based sentiment, we concatenate the topic keyword at the end of the Tweet text to convert it into a text-based classification problem. SemEval comes in data split across years from 2013 to 2017. We evaluate our models on train test splits from each year to assess the temporal stability of our model. We mark yearly evaluation as SemEval 1 and aggregate task evaluation as SemEval 2 in our results.

\paragraph{SocialMediaIE}
Social Media IE~\cite{SocialMediaIE,MDMT,MishraThesisDSTDIE2020} (SMIE) is a collection of datasets specific for evaluation of Information Extraction Systems for Social Media. It consists of datasets of classification and sequence tagging tasks \cite{MDMT}. We utilize the classification tasks from Social Media IE and use them for our evaluation. We use the macro-F1 score for each task. Similar to SemEval this dataset is also released as a set of Tweet IDs and labels, hence we hydrate it using the same approach as SemEval dataset.

\paragraph{TweetEval}
TweetEval \cite{barbieri-etal-2020-tweeteval} was released as a benchmark of classification tasks for Tweets. It consists of anonymized Tweet texts without Tweet Ids. The Tweet text has been anonymized by removing user mentions. This limits us to only use Hashtag-based NTUs for this dataset but we include this dataset to highlight the utility of our approach on this standard benchmark.

\setlength{\tabcolsep}{2pt}
\begin{table*}[]
\centering
\begin{tabular}{@{}lcccrrccc@{}}
\toprule
\textbf{Model} & {\textbf{NTUs}} & {\textbf{Perplexity}} & {\textbf{Topic}} & {\textbf{TweetEval}} & {\textbf{SemEval 1}} & {\textbf{SemEval 2}} & {\textbf{Hashtag}} & {\textbf{SMIE}} \\
& & bits & MAP & mean F1 & mean F1 & mean F1 & Recall@10 & mean F1 \\
\midrule
\textbf{BERT}  & \textbf{-}   & 4.425 & 0.327 & 0.577 & 0.527 & 0.515 & 0.689 & 0.548 \\ 
\midrule
\textbf{\ntulm{}} & \textbf{author} & 4.412 & 0.325 & 0.579 & 0.527 & \textbf{0.548} & 0.693 & 0.548 \\
\textbf{\ntulm{}} & \textbf{Hashtag} & 4.391 & 0.339 & 0.586 & 0.534 & 0.545 & 0.711 & 0.539 \\
\textbf{\ntulm{}} & \textbf{author+Hashtag} & \textbf{4.344} & \textbf{0.343} & \textbf{0.590} & \textbf{0.534} & 0.545 & \textbf{0.720} & \textbf{0.549}                             \\ 
\bottomrule
\end{tabular}
\caption{\ntulm{} compared with BERT (MLM fine-tuned, section \ref{sec:mlm-finetuning}). We report the perplexity, mean average precision (MAP) in Topic, Recall@$10$ in Hashtag Prediction, and mean F1 score in the rest.}
\label{tab:kelm-bert-all}
\end{table*}

\subsection{MLM Fine-tuning}
\label{sec:mlm-finetuning}
We start by fine-tuning the BERT and \ntulm{} models on 1M Tweet data randomly sampled from latest English tweets posted between 2022-06-01 and 2022-06-15. We experiment with training using different contexts. We only consider the inclusion of author and Hashtag contexts as they are the highest coverage contexts across all the datasets. User mentions are few and, in most datasets, they are anonymized. In MLM fine-tuning, we keep all the hyperparameters of \ntulm{} model the same as the BERT baselines.

\subsection{Downstream Task Evaluation}
For each task we feed the unified \ntulm{} embedding $embed_{\ntulm{}}$ into a 2-layer perceptron (MLP) with the final layer being a softmax over possible labels. For topic classification, we use a sigmoid activation for multiple labels. We use the task-specific evaluation to compare the model. We report aggregate improvement on each dataset using the average of metrics for each task in the dataset. Often we report the percentage gains over the BERT model, i.e. $\frac{score_{\ntulm{}} - score_{BERT}}{score_{BERT}} * 100$, this is positive when \ntulm{} is better than BERT. It denotes the percentage \ntulm{} is better or worse than the BERT model. Absolute scores are in table \ref{tab:all-task-performance}. In the experiments of downstream tasks, we keep MLP architectures and hyper-parameters the same for \ntulm{} and baselines.



\section{Evaluation Results}\label{sec:result}

\subsection{Perplexity Experiments}
As highlighted in Table~\ref{tab:kelm-bert-all} we find that the MLM perplexity (lower is better) of all the \ntulm{} models is much better than the perplexity of the BERT-based model. In terms of percentage change, \ntulm{} (author+Hashtag) has about $2\%$ gain in perplexity than the BERT model. This highlights that using contextual information helps improve the MLM task performance. This result is aligned with the findings of LMSOC \cite{kulkarni-etal-2021-lmsoc-approach} that also found that using temporal and geographic context leads to better language modeling. Our work highlights that the graph context of authors and Hashtags encodes additional information which can help in better modeling of the text. We also observe that the Hashtag and author information alone is helpful in lowering the perplexity of the model with Hashtags being more effective. This is also aligned with the usage of Hashtags. Authors on Twitter often use Hashtags to supply topical or community information to a Tweet. Hence, using a Hashtag's graph information improves the model's prediction of the masked words. 

\begin{table*}[]
\centering
\begin{tabular}{@{}ll|rrrrr|r@{}}
\toprule
\textbf{Dataset}       & \textbf{\begin{tabular}[c]{@{}l@{}}Sub-Dataset\\ or Metric\end{tabular}} & \textbf{BERT}                                                   & \textbf{\begin{tabular}[c]{@{}r@{}}\ntulm{}\\ user\end{tabular}} & \textbf{\begin{tabular}[c]{@{}r@{}}\ntulm{}\\ hashtag\end{tabular}} & \textbf{\begin{tabular}[c]{@{}r@{}}\ntulm{}\\ user+hashtag\end{tabular}} & \textbf{\begin{tabular}[c]{@{}r@{}}BERT\\post-concat\end{tabular}}                                                  & \textbf{Best}                               \\
\midrule
\textbf{Topic}         & \textbf{topic}                                                           & 32.65\%                                                         & 32.49\%                                                        & 33.91\%                                                           & {\color[HTML]{4285F4} {\ul 34.32\%}}                                   & {\color[HTML]{EA4335} \textbf{38.76\%}}                         & \cellcolor[HTML]{B7E1CD}BERT-post-concat    \\
\midrule
\textbf{Hashtag}       & \textbf{recall@10}                                                       & 68.88\%                                                         & 69.26\%                                                        & 71.09\%                                                           & {\color[HTML]{4285F4} {\ul 71.99\%}}                                   & {\color[HTML]{EA4335} \textbf{72.23\%}}                         & \cellcolor[HTML]{B7E1CD}BERT-post-concat    \\
\midrule
\textbf{TweetEval}     & \textbf{emoji}                                                           & 18.02\%                                 & 18.10\%                                & 18.44\%                                   & {\color[HTML]{4285F4} {\ul 18.55\%}}           & {\color[HTML]{EA4335} \textbf{19.07\%}} & \cellcolor[HTML]{B7E1CD}BERT-post-concat    \\
\textbf{TweetEval}     & \textbf{emotion}                                                         & {\color[HTML]{EA4335} \textbf{67.70\%}} & {\color[HTML]{4285F4} {\ul 67.65\%}}   & 66.61\%                                   & 67.31\%                                        & 67.60\%                                 & \cellcolor[HTML]{F4C7C3}BERT                \\
\textbf{TweetEval}     & \textbf{hate}                                                            & {\color[HTML]{EA4335} \textbf{59.50\%}} & {\color[HTML]{4285F4} {\ul 58.59\%}}   & 56.87\%                                   & 58.16\%                                        & 57.83\%                                 & \cellcolor[HTML]{F4C7C3}BERT                \\
\textbf{TweetEval}     & \textbf{irony}                                                           & 60.37\%                                 & 62.03\%                                & {\color[HTML]{EA4335} \textbf{66.67\%}}   & {\color[HTML]{4285F4} {\ul 66.17\%}}           & 58.88\%                                 & \cellcolor[HTML]{FCE8B2}\ntulm{} (hashtag)      \\
\textbf{TweetEval}     & \textbf{offensive}                                                       & 72.51\%                                 & 72.73\%                                & {\color[HTML]{EA4335} \textbf{73.71\%}}   & {\color[HTML]{4285F4} {\ul 73.63\%}}           & 71.52\%                                 & \cellcolor[HTML]{FCE8B2}\ntulm{} (hashtag)      \\
\textbf{TweetEval}     & \textbf{sentiment}                                                       & 60.66\%                                 & {\color[HTML]{4285F4} {\ul 61.40\%}}   & 60.66\%                                   & {\color[HTML]{EA4335} \textbf{61.43\%}}        & 58.65\%                                 & \cellcolor[HTML]{FCE8B2}\ntulm{} (user+hashtag) \\
\textbf{TweetEval}     & \textbf{stance}                                                          & 64.88\%                                 & 65.11\%                                & {\color[HTML]{4285F4} {\ul 67.48\%}}      & {\color[HTML]{EA4335} \textbf{67.56\%}}        & 66.89\%                                 & \cellcolor[HTML]{FCE8B2}\ntulm{} (user+hashtag) \\
\midrule
\textbf{SemEval 1}       & \textbf{2013-A}                                                          & 67.75\%                                                         & 67.61\%                                                        & {\color[HTML]{4285F4} {\ul 67.94\%}}                              & {\color[HTML]{EA4335} \textbf{68.38\%}}                                & 67.54\%                                                         & \cellcolor[HTML]{FCE8B2}\ntulm{} (user+hashtag) \\
\textbf{SemEval 1}       & \textbf{2014-A}                                                          & 26.80\%                                                         & 26.06\%                                                        & {\color[HTML]{EA4335} \textbf{27.96\%}}                           & 26.91\%                                                                & {\color[HTML]{4285F4} {\ul 27.48\%}}                            & \cellcolor[HTML]{FCE8B2}\ntulm{} (hashtag)      \\
\textbf{SemEval 1}       & \textbf{2015-A}                                                          & 53.70\%                                                         & 53.73\%                                                        & {\color[HTML]{EA4335} \textbf{54.63\%}}                           & {\color[HTML]{4285F4} {\ul 54.63\%}}                                   & 53.31\%                                                         & \cellcolor[HTML]{FCE8B2}\ntulm{} (hashtag)      \\
\textbf{SemEval 1}       & \textbf{2015-BD}                                                         & 41.17\%                                                         & {\color[HTML]{4285F4} {\ul 41.36\%}}                           & 40.45\%                                                           & 41.08\%                                                                & {\color[HTML]{EA4335} \textbf{41.61\%}}                         & \cellcolor[HTML]{B7E1CD}BERT-post-concat    \\
\textbf{SemEval 1}       & \textbf{2016-A}                                                          & 51.38\%                                                         & 52.52\%                                                        & {\color[HTML]{4285F4} {\ul 53.01\%}}                              & {\color[HTML]{EA4335} \textbf{53.70\%}}                                & 51.50\%                                                         & \cellcolor[HTML]{FCE8B2}\ntulm{} (user+hashtag) \\
\textbf{SemEval 1}       & \textbf{2016-BD}                                                         & 92.60\%                                                         & {\color[HTML]{4285F4} {\ul 92.65\%}}                           & {\color[HTML]{EA4335} \textbf{92.71\%}}                           & 92.56\%                                                                & 92.58\%                                                         & \cellcolor[HTML]{FCE8B2}\ntulm{} (hashtag)      \\
\textbf{SemEval 1}       & \textbf{2016-CE}                                                         & 35.58\%                                                         & 35.20\%                                                        & {\color[HTML]{EA4335} \textbf{36.86\%}}                           & {\color[HTML]{4285F4} {\ul 36.74\%}}                                   & 35.25\%                                                         & \cellcolor[HTML]{FCE8B2}\ntulm{} (hashtag)      \\
\midrule
\textbf{SemEval 2}       & \textbf{task-A}                                                          & 48.02\%                                                         & 47.91\%                                                        & 47.54\%                                                           & {\color[HTML]{EA4335} \textbf{49.72\%}}                                & {\color[HTML]{4285F4} {\ul 48.71\%}}                            & \cellcolor[HTML]{FCE8B2}\ntulm{} (user+hashtag) \\
\textbf{SemEval 2}       & \textbf{task-BD}                                                         & 71.56\%                                                         & 71.92\%                                                        & {\color[HTML]{4285F4} {\ul 71.95\%}}                              & {\color[HTML]{EA4335} \textbf{72.59\%}}                                & 71.33\%                                                         & \cellcolor[HTML]{FCE8B2}\ntulm{} (user+hashtag) \\
\textbf{SemEval 2}       & \textbf{task-CE}                                                         & {\color[HTML]{4285F4} {\ul 34.83\%}}                            & 34.69\%                                                        & {\color[HTML]{EA4335} \textbf{34.95\%}}                           & 33.92\%                                                                & 34.71\%                                                         & \cellcolor[HTML]{FCE8B2}\ntulm{} (hashtag)      \\
\midrule
\textbf{SMIE} & \textbf{abusive 1}                                                       & {\color[HTML]{4285F4} {\ul 55.84\%}}                            & {\color[HTML]{EA4335} \textbf{56.27\%}}                        & 55.08\%                                                           & 55.69\%                                                                & 54.04\%                                                         & \cellcolor[HTML]{FCE8B2}\ntulm{} (user)         \\
\textbf{SMIE} & \textbf{abusive 2}                                                       & {\color[HTML]{4285F4} {\ul 47.36\%}}                            & 47.04\%                                                        & 44.82\%                                                           & {\color[HTML]{EA4335} \textbf{48.00\%}}                                & 37.01\%                                                         & \cellcolor[HTML]{FCE8B2}\ntulm{} (user+hashtag) \\
\textbf{SMIE} & \textbf{sentiment 1}                                                     & {\color[HTML]{EA4335} \textbf{76.01\%}}                         & 74.52\%                                                        & 74.73\%                                                           & {\color[HTML]{4285F4} {\ul 75.14\%}}                                   & 73.93\%                                                         & \cellcolor[HTML]{F4C7C3}BERT                \\
\textbf{SMIE} & \textbf{sentiment 2}                                                     & 61.86\%                                                         & {\color[HTML]{EA4335} \textbf{62.20\%}}                        & 61.70\%                                                           & {\color[HTML]{4285F4} {\ul 61.92\%}}                                   & 61.61\%                                                         & \cellcolor[HTML]{FCE8B2}\ntulm{} (user)         \\
\textbf{SMIE} & \textbf{sentiment 3}                                                     & 58.69\%                                                         & {\color[HTML]{4285F4} {\ul 58.73\%}}                           & {\color[HTML]{EA4335} \textbf{58.80\%}}                           & 58.43\%                                                                & 58.70\%                                                         & \cellcolor[HTML]{FCE8B2}\ntulm{} (hashtag)      \\
\textbf{SMIE} & \textbf{sentiment 4}                                                     & 53.78\%                                                         & 54.75\%                                                        & 55.68\%                                                           & {\color[HTML]{4285F4} {\ul 56.48\%}}                                   & {\color[HTML]{EA4335} \textbf{57.23\%}}                         & \cellcolor[HTML]{B7E1CD}BERT-post-concat    \\
\textbf{SMIE} & \textbf{sentiment 5}                                                     & {\color[HTML]{EA4335} \textbf{60.22\%}}                         & 59.65\%                                                        & {\color[HTML]{4285F4} {\ul 59.86\%}}                              & 59.77\%                                                                & 57.99\%                                                         & \cellcolor[HTML]{F4C7C3}BERT                \\
\textbf{SMIE} & \textbf{sentiment 6}                                                     & 59.66\%                                                         & 59.58\%                                                        & {\color[HTML]{EA4335} \textbf{60.15\%}}                           & {\color[HTML]{4285F4} {\ul 59.81\%}}                                   & 59.43\%                                                         & \cellcolor[HTML]{FCE8B2}\ntulm{} (hashtag)      \\
\textbf{SMIE} & \textbf{uncertainity 1}                                                  & 55.37\%                                                         & 55.81\%                                                        & 51.52\%                                                           & {\color[HTML]{4285F4} {\ul 56.00\%}}                                   & {\color[HTML]{EA4335} \textbf{57.14\%}}                         & \cellcolor[HTML]{B7E1CD}BERT-post-concat    \\
\textbf{SMIE} & \textbf{uncertainity 2}                                                  & 19.03\%                                                         & {\color[HTML]{4285F4} {\ul 19.05\%}}                           & 16.80\%                                                           & 17.63\%                                                                & {\color[HTML]{EA4335} \textbf{19.11\%}}                         & \cellcolor[HTML]{B7E1CD}BERT-post-concat   \\
\bottomrule
\end{tabular}
\caption{
Absolute metrics across all tasks and their subtasks. {\color[HTML]{EA4335} \textbf{Best score}} and {\color[HTML]{4285F4} {\ul Second best score}}. SMIE=SocialMediaIE, BERTC=BERT-post-concat with user+Hashtag NTUs, BERT=BERT (MLM fine-tuned, section \ref{sec:mlm-finetuning}).
}\label{tab:all-task-performance}
\end{table*}

\begin{table*}[]
\centering
\begin{tabular}{@{}l|rr|rr|rr@{}}
\toprule
\textbf{Dataset}       & \multicolumn{2}{c}{\textbf{Overall}} & \multicolumn{2}{c}{\textbf{Overlap}} & \multicolumn{2}{c}{\textbf{Non-Overlap}} \\ \midrule
                      & \textbf{\ntulm{}}    & \textbf{BERTC}    & \textbf{\ntulm{}}    & \textbf{BERTC}    & \textbf{\ntulm{}}      & \textbf{BERTC}      \\
\midrule
\textbf{TweetEval}     & 2.27\%           & -0.80\%           & 2.73\%           & -3.33\%           & 0.31\%             & 0.65\%              \\
\textbf{SemEval 1}     & 1.36\%           & 0.08\%            & 2.59\%           & 0.21\%            & 0.65\%             & 0.02\%              \\
\textbf{SemEval 2}     & 5.93\%           & 0.22\%            & -0.07\%          & 0.58\%            & 2.62\%             & 0.07\%              \\
\textbf{SocialMediaIE} & 0.20\%           & -2.12\%           & -0.27\%          & -4.12\%           & 1.98\%             & -22.22\%            \\
\textbf{Hashtag}       & 4.51\%           & 4.87\%            & 5.61\%           & 7.46\%            & 1.01\%             & -3.37\%             \\
\textbf{Topic}         & 5.10\%           & 18.72\%           & 6.92\%           & 34.72\%           & 0.71\%             & -4.17\%             \\ \bottomrule
\end{tabular}%
\caption{\% improvement over BERT (MLM fine-tuned see section \ref{sec:mlm-finetuning}) by using user+Hashtag NTUs in \ntulm{} versus BERT-post-concat (BERTC) across datasets, and split across overlapping and non-overlapping subsets.}
\label{tab:kelm-bert-post-concat-all}
\end{table*}

\subsection{Downstream Classification}
Now we look at how the \ntulm{} model performs across various downstream tasks. As highlighted in Table~\ref{tab:kelm-bert-all} (detailed numbers in Table~\ref{tab:all-task-performance}), we see that enriching text with NTU information from author+Hashtag always leads to significant performance improvement over BERT fine-tuned using MLM pre-training on the same dataset as \ntulm{} as explained in section \ref{sec:mlm-finetuning}. Specifically, the author+Hashtag \ntulm{} model is 5\% better than BERT on Topic prediction, 2\% better on TweetEval, 6\% better on SemEval 1, 4.5\% better on SemEval 2, and 0.2\% better on SocialMediaIE.  

\begin{figure}
    \centering
    \includegraphics[width=\linewidth]{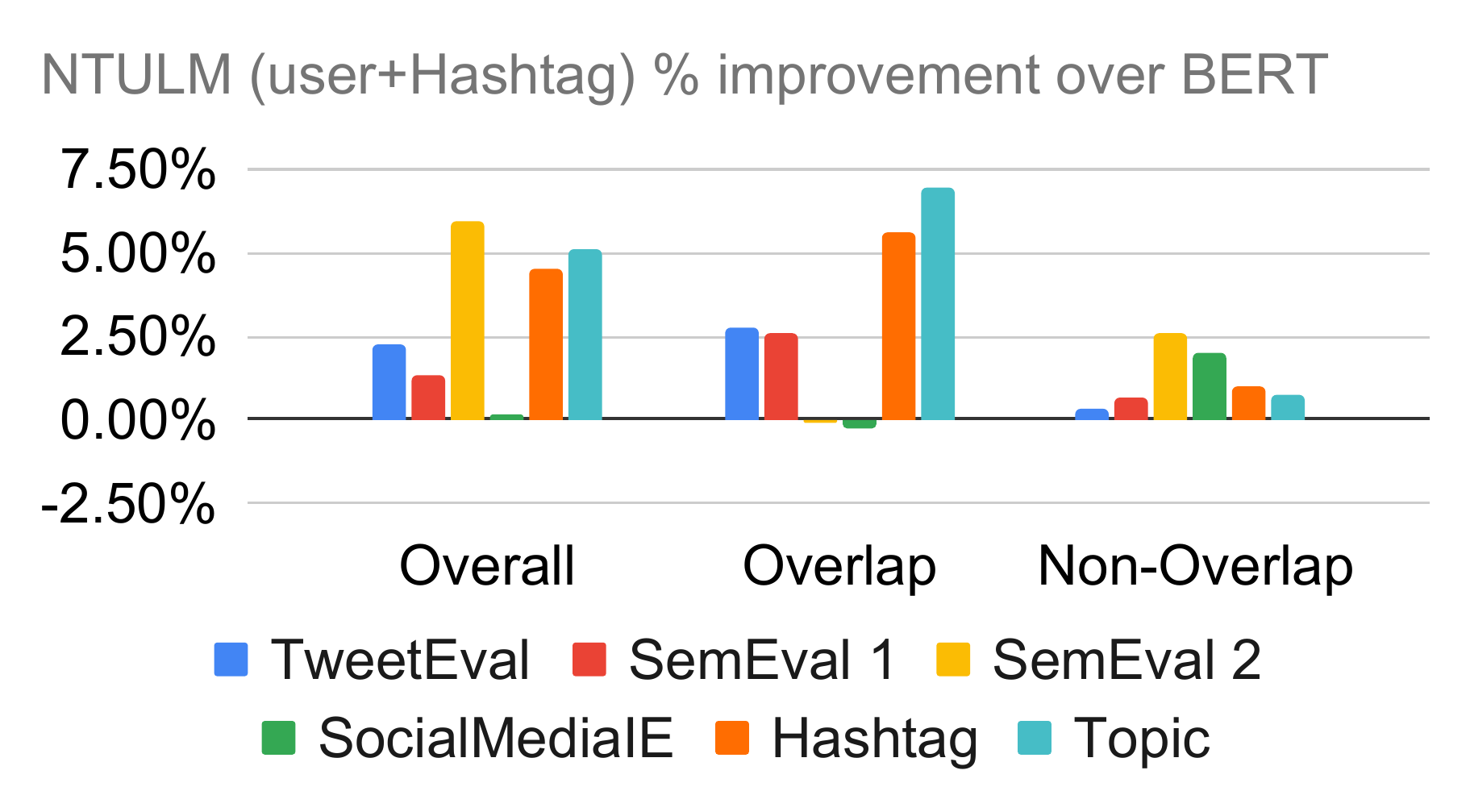}
    \caption{\ntulm{} versus BERT (MLM fine-tuned see section \ref{sec:mlm-finetuning}) on Tweets with and without NTU overlap with NTU embeddings. See Table~\ref{tab:kelm-bert-post-concat-all} for details.}
    \label{fig:all-task-overlap}
    \vspace{-3pt}
\end{figure}

\begin{figure}
    \centering
    \includegraphics[width=\linewidth]{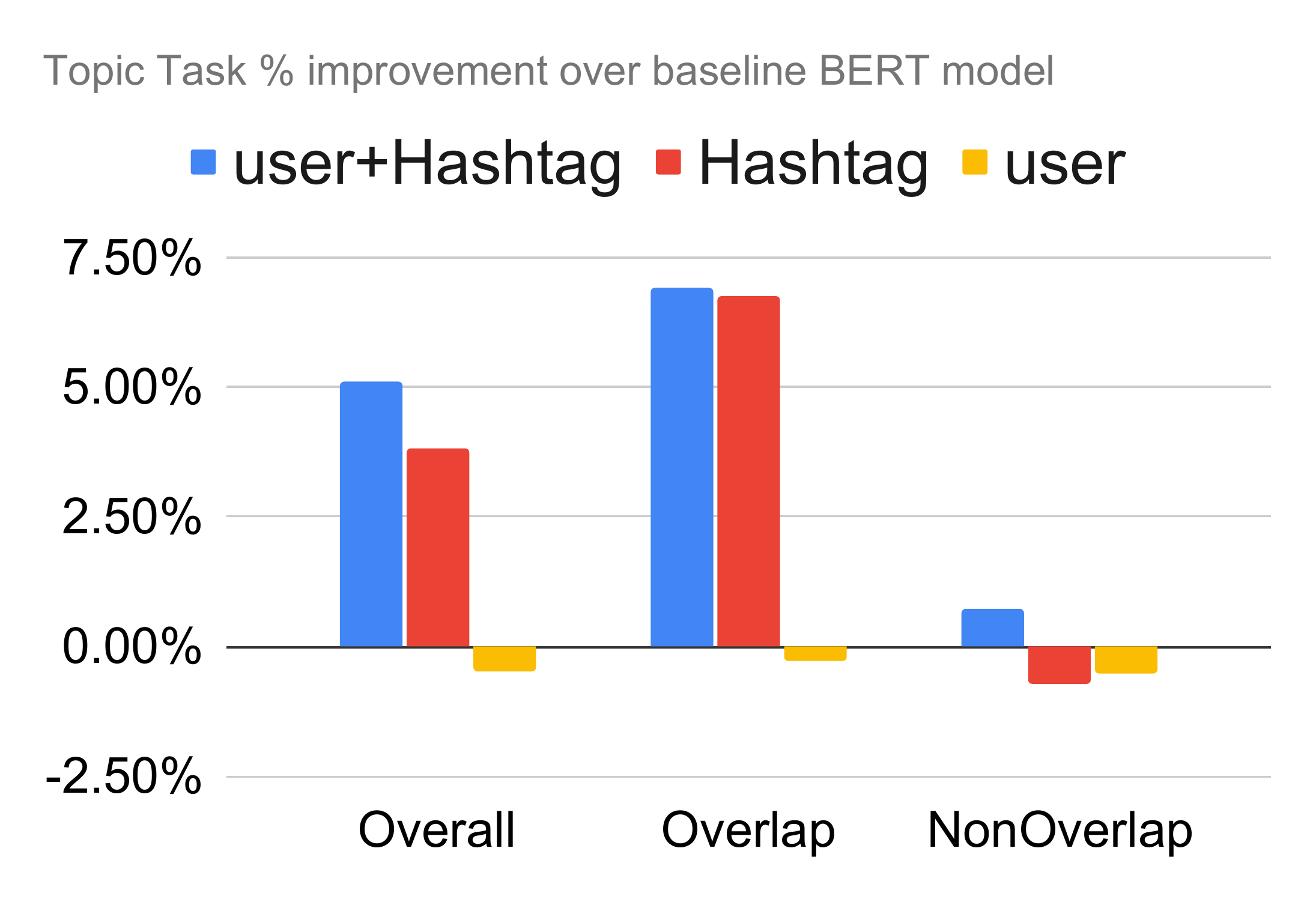}
    \caption{Performance on Tweets with and without NTU overlap with NTU embeddings on Topic prediction task. BERT is MLM fine-tuned see section \ref{sec:mlm-finetuning}.}
    \label{fig:topic-task-overlap}
    \vspace{-3pt}
\end{figure}

Furthermore, we assess how the model's performance changes compared to BERT for Tweets which have NTUs overlapping (Overlap) with our NTU embeddings versus those which do not have Tweets overlapping with the NTU embeddings (Non-overlap). Our focus here is that for Tweets with NTU overlap we should see significant improvement wherease for Tweets without NTU overlap we should not change our performance compared to BERT as we are back to the text-only setting. As highlighted in Figure~\ref{fig:all-task-overlap} and Figure~\ref{fig:topic-task-overlap} we see that the improvement over BERT on the overlap case is higher than the overall improvement for the author+Hashtag \ntulm{} across most tasks. Furthermore, in the no-overlap case, we do not see any significant loss in performance, in fact author+Hashtag is slightly better compared to BERT (0.7\%). This highlights that the NTU contexts are really helping in the downstream tasks whenever the NTUs are available.

\subsection{Case-study: Concatenation vs Attention}

Next, we consider the setting of concatenating the NTU embeddings to BERT embeddings. This is a simple setting where the language model is not able to generate a Text specific embedding based on NTUs. This is a simple baseline which is often adopted when integrating signals from multiple sources. We name this model BERT-post-concat and compare it with our best model \ntulm{} (author+Hashtag). Here again we compare these models against the BERT model which only uses text and was was MLM fine-tuned as explained in section \ref{sec:mlm-finetuning}. 

\begin{figure}
    \centering
    \includegraphics[width=0.9\linewidth]{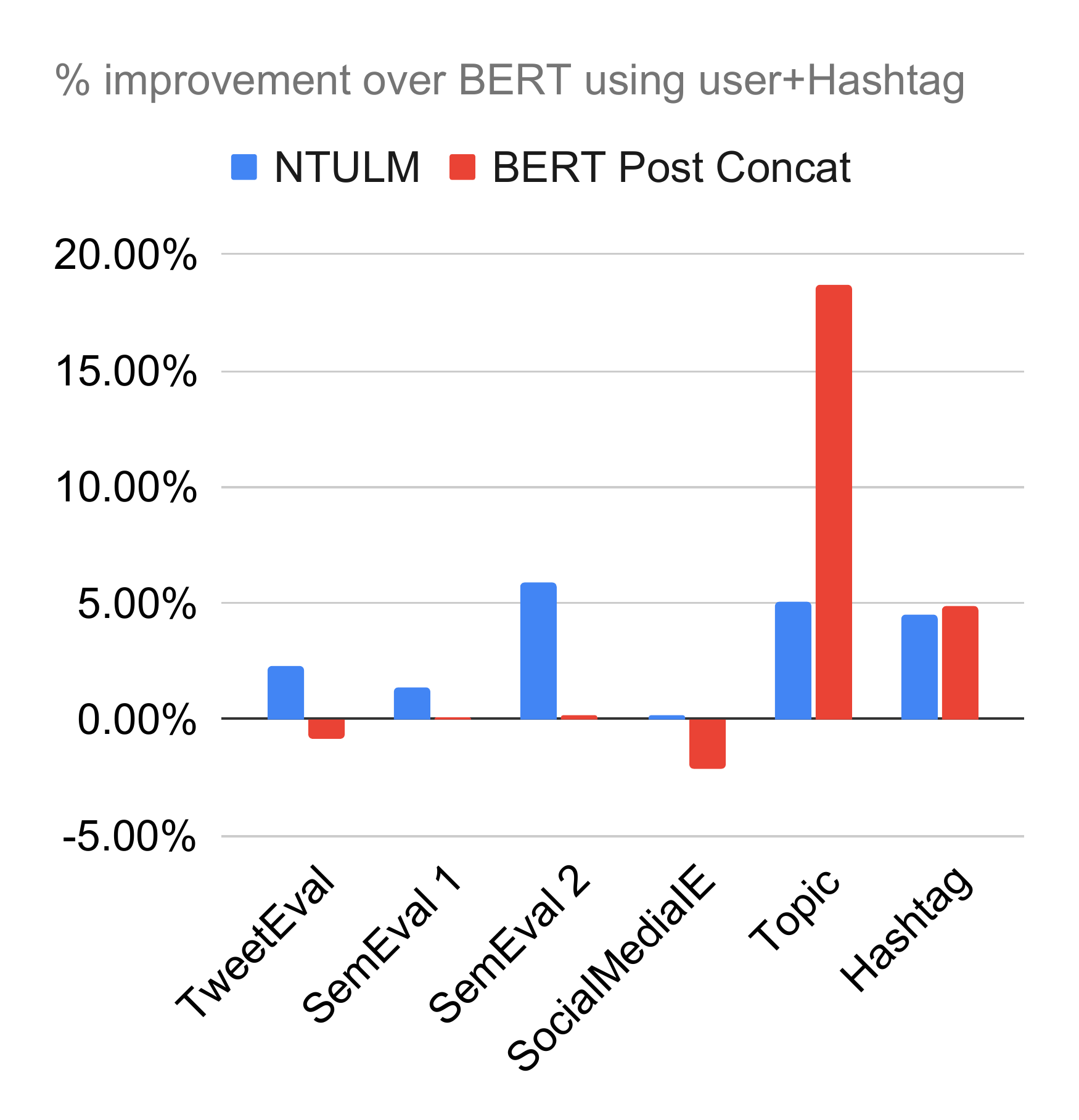}
    \vspace{-6pt}
    \caption{\ntulm{} versus BERT-post-concat as measured in improvement over BERT (MLM fine-tuned see section \ref{sec:mlm-finetuning}) across tasks.}
    \vspace{-3pt}
    \label{fig:bert-concat}
\end{figure}

Figure~\ref{fig:bert-concat} (detailed numbers in Table~\ref{tab:all-task-performance}) highlights that using the \ntulm{} approach is much better than BERT-post-concat for most tasks, except for topic and Hashtag prediction. For Hashtag dataset \ntulm{} is only 0.34\% worse in relative performance compared to BERT-post-concat. However, in the topic prediction task \ntulm{} is -11.5\% worse. We hypothesize that the improved performance of BERT-post-concat is a result of the direct relevance of Hashtag embeddings to the dowstream task of Topic and Hashtag relevance as \ntulm{}'s frozen embedding dilutes this information. We confirm this by inspecting the performance (see Table~\ref{tab:kelm-bert-post-concat-all}) of BERT-post-concat on the overlapping and non-overlapping slices of the data, where BERT post-concat is better than \ntulm{} on the overlapping slice of the data but is worse than \ntulm{} and even BERT on the non-overlapping slice. This highlights that BERT post-concat is overfitting to the NTU signal in the data which is not the case with \ntulm{}. We reason that fine-tuning \ntulm{} for the downstream task may address this issue and plan to explore this in a future work given that the focus of this work is to generate high quality general purpose Tweet embeddings. Furthermore, for TweetEval and SocialMediaIE BERT post-concat performs even worse than BERT. This can be attributed to the indirect relevance of author and Hashtag identity to the downstream tasks in these datasets which the BERT-post-concat cannot capture. 

%% file: 05_relatedwork.tex
\section{Related Work}\label{sec:relatedwork}

\paragraph{Knowledge Graph and Language Models:}
Previous work has investigated language models with knowledge graphs. KI-BERT~\cite{faldu2021ki} extracts and computes the embedding of concepts and ambiguous entities from text and appends them to the end of the sentence to enrich a language model. K-BERT~\cite{liu2020k} uses an external knowledge graph to build a sentence tree and integrates the knowledge graph before the embedding layer of BERT. KEPLER~\cite{wang2021kepler} incorporates knowledge embedding of text entities as an auxiliary objective alongside the traditional MLM objective for BERT. While these models have shown improvements on some domain-specific tasks, they only consider the textual entities from the text itself, which limits their performance in modeling language with rich contextual information (e.g. social networks). Different from existing works, the \ntulm{} framework can incorporate the contextual information of multi-type non-textual units and therefore has a better performance in understanding contexts. There are also some existing works that use social contexts to enrich the language model, such as LMSOC~\cite{kulkarni-etal-2021-lmsoc-approach}. However, instead of considering the non-text units such as author, Hashtag, URL, and mention, LMSOC only considers time and location. In addition, LMSOC only supports incorporating one type of social context, which limits its performance on texts with rich contexts. 

\paragraph{Representation Learning of Social Graph:}
Learning the representation of social entities such as tweets and users has been a popular research topic over the past few years. InfoVGAE~\cite{li2022info} constructs a bipartite heterogeneous graph and designs an orthogonal latent space to learn explainable user and tweet embeddings. In kNN-Embed~\cite{el2022knn}, a bi-partite Twitter follow graph is embedded for account suggestion. TIMME~\cite{xiao2020timme} uses multi-task learning of link prediction and entity classification to jointly learn the representation of tweets. SEM~\cite{pougue2022learning} creates a topical Twitter agreement graph and embeds nodes via a random-walk approach to detect user stances on given topics. \cite{zhang2022kdd} proposes a second-order continuous GNN to improve the social network embeddings. Most of these models do not consider textual information of social graph. Only the interaction data is applied to learn the representation of social entities, which limits their performance on downstream tasks. 

\paragraph{Language Model for Social Networks:}
Many existing works have explored the training of language models in the social network domain. Tweet2vec~\cite{vosoughi2016tweet2vec} proposes a character-level CNN-LSTM encoder-decoder to improve the tweet embeddings. DICE~\cite{naseem2019dice} leverages contextual text to address polysemy and improve the tweet embedding quality. TweeTIME~\cite{tabassum2016tweetime} proposes a minimally supervised method to address the time recognition problem from Twitter texts. TweetBERT~\cite{qudar2020tweetbert} models are trained on the domain-specific data of tweet texts and outperform traditional BERT models. However, most of these language model does not take NTUs into consideration and cannot benefit from the interaction and engagement data.

%% file: 06_conclusion.tex
\section{Limitations}
\label{sec:limitations}
One major limitation of our work is the averaging of heterogenous embeddings. This approach works because the embeddings trained using TransE lie in the same space but is less expressive as we are not including explicit information around which type of NTU an embedding is coming from. In future we plan to address this by including type specific embedding transformation before doing an averaging. However, given the results, this naive averaging of user+Hashtag still works well across tasks it shows the utility of our approach.
Next, our training data is relatively small and less diverse with only 1M Tweets as budgetary and computational constraints influenced our experimental setup. In this paper, our goal has been to demonstrate the effectiveness of our approach paving the way for future work that scales up the training and uses a much larger and more diverse dataset.  
Finally, our results are on English specific datasets and models. While the utilization of NTU embeddings make our approach language agnostic, in future we plan to demonstrate its impact across multiple languages. 

\section{Conclusion}
\label{sec:conclude}
In this paper we introduced NTU enriched Language Model (\ntulm{}), a method of enriching a pretrained BERT model by adding graph embeddings of non-textual units. We experimentally demonstrate that including NTU representations alongside text yields superior representations vs a text-only language model. On several downstream tasks, we show significant improvment using \ntulm{} representations compared to BERT-based sentence embeddings.

%% file: 07_appendix.tex
\onecolumn
\section{Appendix: Dataset Statistics}

Here we provide the statistics of our datasets for downstream evaluation experiments in Table~\ref{tab:downstream-data-stats}.

\setlength{\tabcolsep}{2pt}
\begin{longtable}[!htbp]{@{}rrr|r|rrrr|rrrr@{}}
\toprule
\textbf{}        & \textbf{}            & \textbf{}      & \textbf{}             &
\multicolumn{4}{c}{\textbf{Hashtag}}                                                                             & \multicolumn{4}{c}{\textbf{User}}                                                                                \\* 
\textbf{dataset} & \textbf{task} & \textbf{split} & \textbf{Tweets} & \textbf{NTUs} & \textbf{\textgreater 1 NTUs} & \textbf{\textgreater 1 $\in$ E} & \textbf{ $\in$ E} & \textbf{NTUs} & \textbf{\textgreater 1 NTUs} & \textbf{\textgreater 1 $\in$ E} & \textbf{ $\in$ E} \\*
\midrule
\endfirsthead
\multicolumn{12}{c}%
{{\bfseries Table \thetable\ continued from previous page}} \\
\toprule
\textbf{}        & \textbf{}            & \textbf{}      & \textbf{}             & \multicolumn{4}{c}{\textbf{Hashtag}}                                                                             & \multicolumn{4}{c}{\textbf{User}}                                                                                \\* 
\textbf{dataset} & \textbf{task} & \textbf{split} & \textbf{Tweets} & \textbf{NTUs} & \textbf{\textgreater 1 NTUs} & \textbf{\textgreater 1 $\in$ E} & \textbf{ $\in$ E} & \textbf{NTUs} & \textbf{\textgreater 1 NTUs} & \textbf{\textgreater 1 $\in$ E} & \textbf{ $\in$ E} \\*
\midrule
\endhead
\bottomrule
\endfoot
\endlastfoot
%
TweetEval        & emoji                & train          & 45,000                & 28,251               & 46.37\%                      & 42.82\%                             & 92\%                 & 0                    & 0.00\%                       & 0.00\%                              & 0\%                  \\
TweetEval        & emoji                & test           & 50,000                & 30,989               & 43.10\%                      & 39.68\%                             & 92\%                 & 0                    & 0.00\%                       & 0.00\%                              & 0\%                  \\
TweetEval        & emotion              & train          & 3,257                 & 1,652                & 43.94\%                      & 43.14\%                             & 98\%                 & 0                    & 0.00\%                       & 0.00\%                              & 0\%                  \\
TweetEval        & emotion              & test           & 1,421                 & 1,071                & 47.29\%                      & 46.94\%                             & 99\%                 & 0                    & 0.00\%                       & 0.00\%                              & 0\%                  \\
TweetEval        & hate                 & train          & 9,000                 & 2,375                & 25.69\%                      & 25.10\%                             & 98\%                 & 0                    & 0.00\%                       & 0.00\%                              & 0\%                  \\
TweetEval        & hate                 & test           & 2,970                 & 1,615                & 50.20\%                      & 49.70\%                             & 99\%                 & 0                    & 0.00\%                       & 0.00\%                              & 0\%                  \\
TweetEval        & irony                & train          & 2,862                 & 2,132                & 38.36\%                      & 36.09\%                             & 94\%                 & 0                    & 0.00\%                       & 0.00\%                              & 0\%                  \\
TweetEval        & irony                & test           & 784                   & 857                  & 72.19\%                      & 71.30\%                             & 99\%                 & 0                    & 0.00\%                       & 0.00\%                              & 0\%                  \\
TweetEval        & offensive            & train          & 11,916                & 1,937                & 14.40\%                      & 14.10\%                             & 98\%                 & 0                    & 0.00\%                       & 0.00\%                              & 0\%                  \\
TweetEval        & offensive            & test           & 860                   & 1,276                & 73.26\%                      & 71.28\%                             & 97\%                 & 0                    & 0.00\%                       & 0.00\%                              & 0\%                  \\
TweetEval        & sentiment            & train          & 45,615                & 6,956                & 18.35\%                      & 16.63\%                             & 91\%                 & 0                    & 0.00\%                       & 0.00\%                              & 0\%                  \\
TweetEval        & sentiment            & test           & 12,284                & 3,933                & 39.14\%                      & 37.63\%                             & 96\%                 & 0                    & 0.00\%                       & 0.00\%                              & 0\%                  \\
TweetEval        & stance 1             & train          & 587                   & 455                  & 95.91\%                      & 95.91\%                             & 100\%                & 0                    & 0.00\%                       & 0.00\%                              & 0\%                  \\
TweetEval        & stance 1             & test           & 280                   & 277                  & 100.00\%                     & 100.00\%                            & 100\%                & 0                    & 0.00\%                       & 0.00\%                              & 0\%                  \\
TweetEval        & stance 2             & train          & 461                   & 423                  & 100.00\%                     & 100.00\%                            & 100\%                & 0                    & 0.00\%                       & 0.00\%                              & 0\%                  \\
TweetEval        & stance 2             & test           & 220                   & 251                  & 100.00\%                     & 100.00\%                            & 100\%                & 0                    & 0.00\%                       & 0.00\%                              & 0\%                  \\
TweetEval        & stance 3             & train          & 355                   & 416                  & 100.00\%                     & 100.00\%                            & 100\%                & 0                    & 0.00\%                       & 0.00\%                              & 0\%                  \\
TweetEval        & stance 3             & test           & 169                   & 201                  & 100.00\%                     & 100.00\%                            & 100\%                & 0                    & 0.00\%                       & 0.00\%                              & 0\%                  \\
TweetEval        & stance 4             & train          & 597                   & 353                  & 100.00\%                     & 100.00\%                            & 100\%                & 0                    & 0.00\%                       & 0.00\%                              & 0\%                  \\
TweetEval        & stance 4             & test           & 285                   & 198                  & 100.00\%                     & 100.00\%                            & 100\%                & 0                    & 0.00\%                       & 0.00\%                              & 0\%                  \\
TweetEval        & stance 5             & train          & 620                   & 407                  & 97.10\%                      & 97.10\%                             & 100\%                & 0                    & 0.00\%                       & 0.00\%                              & 0\%                  \\
TweetEval        & stance 5             & test           & 295                   & 201                  & 100.00\%                     & 100.00\%                            & 100\%                & 0                    & 0.00\%                       & 0.00\%                              & 0\%                  \\
\midrule
Topic            & topic                & train          & 100,000               & 57,873               & 38.59\%                      & 38.25\%                             & 99\%                 & 89,091               & 100.00\%                     & 14.05\%                             & 14\%                 \\
Topic            & topic                & test           & 20,000                & 17,122               & 38.26\%                      & 37.90\%                             & 99\%                 & 19,006               & 100.00\%                     & 14.19\%                             & 14\%                 \\
\midrule
Hashtag    & hashtag              & train          & 899,606               & 282,603              & 70.92\%                      & 70.49\%                             & 99\%                 & 392,751              & 100.00\%                     & 9.76\%                              & 10\%                 \\
Hashtag    & hashtag              & test           & 100,372               & 64,939               & 70.64\%                      & 70.23\%                             & 99\%                 & 67,903               & 100.00\%                     & 9.65\%                              & 10\%                 \\
\midrule
SemEval          & 2013-A               & train          & 7,110                 & 1,599                & 20.03\%                      & 17.86\%                             & 89\%                 & 9,069                & 100.00\%                     & 23.52\%                             & 24\%                 \\
SemEval          & 2013-A               & test           & 2,284                 & 573                  & 20.53\%                      & 18.13\%                             & 88\%                 & 2,814                & 100.00\%                     & 24.87\%                             & 25\%                 \\
SemEval          & 2014-A               & train          & 30                    & 14                   & 100.00\%                     & 96.67\%                             & 97\%                 & 49                   & 100.00\%                     & 16.67\%                             & 17\%                 \\
SemEval          & 2014-A               & test           & 1,253                 & 254                  & 16.12\%                      & 13.89\%                             & 86\%                 & 1,563                & 100.00\%                     & 26.18\%                             & 26\%                 \\
SemEval          & 2015-A               & train          & 318                   & 71                   & 22.33\%                      & 20.75\%                             & 93\%                 & 412                  & 100.00\%                     & 20.75\%                             & 21\%                 \\
SemEval          & 2015-A               & test           & 1,461                 & 329                  & 20.88\%                      & 19.37\%                             & 93\%                 & 1,887                & 100.00\%                     & 21.15\%                             & 21\%                 \\
SemEval          & 2015-BD              & train          & 316                   & 71                   & 22.47\%                      & 20.89\%                             & 93\%                 & 408                  & 100.00\%                     & 20.57\%                             & 21\%                 \\
SemEval          & 2015-BD              & test           & 1,454                 & 333                  & 21.05\%                      & 19.46\%                             & 92\%                 & 1,887                & 100.00\%                     & 21.18\%                             & 21\%                 \\
SemEval          & 2016-A               & train          & 6,180                 & 1,230                & 17.52\%                      & 15.95\%                             & 91\%                 & 7,775                & 100.00\%                     & 21.13\%                             & 21\%                 \\
SemEval          & 2016-A               & test           & 12,754                & 1,932                & 19.53\%                      & 17.88\%                             & 92\%                 & 14,822               & 100.00\%                     & 20.31\%                             & 20\%                 \\
SemEval          & 2016-BD              & train          & 4,404                 & 977                  & 18.35\%                      & 16.53\%                             & 90\%                 & 5,586                & 100.00\%                     & 22.48\%                             & 22\%                 \\
SemEval          & 2016-BD              & test           & 6,494                 & 1,079                & 19.16\%                      & 17.51\%                             & 91\%                 & 7,776                & 100.00\%                     & 21.40\%                             & 21\%                 \\
SemEval          & 2016-CE              & train          & 6,180                 & 1,230                & 17.52\%                      & 15.95\%                             & 91\%                 & 7,775                & 100.00\%                     & 21.13\%                             & 21\%                 \\
SemEval          & 2016-CE              & test           & 12,754                & 1,932                & 19.53\%                      & 17.88\%                             & 92\%                 & 14,822               & 100.00\%                     & 20.31\%                             & 20\%                 \\
SemEval          & task-A               & train          & 31,019                & 5,296                & 19.32\%                      & 17.50\%                             & 91\%                 & 37,154               & 100.00\%                     & 21.82\%                             & 22\%                 \\
SemEval          & task-A               & test           & 4,609                 & 1,483                & 28.77\%                      & 26.93\%                             & 94\%                 & 5,919                & 100.00\%                     & 17.40\%                             & 17\%                 \\
SemEval          & task-BD              & train          & 11,675                & 2,143                & 19.08\%                      & 17.40\%                             & 91\%                 & 14,245               & 100.00\%                     & 21.72\%                             & 22\%                 \\
SemEval          & task-BD              & test           & 2,324                 & 656                  & 26.25\%                      & 24.44\%                             & 93\%                 & 3,234                & 100.00\%                     & 16.70\%                             & 17\%                 \\
SemEval          & task-CE              & train          & 18,887                & 3,009                & 18.88\%                      & 17.25\%                             & 91\%                 & 22,223               & 100.00\%                     & 20.59\%                             & 21\%                 \\
SemEval          & task-CE              & test           & 4,606                 & 1,485                & 28.90\%                      & 27.05\%                             & 94\%                 & 5,914                & 100.00\%                     & 17.41\%                             & 17\%                 \\
\midrule
SMIE    & abusive 1            & train          & 32,997                & 11,177               & 30.06\%                      & 27.98\%                             & 93\%                 & 48,619               & 100.00\%                     & 23.81\%                             & 24\%                 \\
SMIE    & abusive 1            & test           & 9,070                 & 3,619                & 29.49\%                      & 27.67\%                             & 94\%                 & 14,272               & 100.00\%                     & 23.24\%                             & 23\%                 \\
SMIE    & abusive 2            & train          & 8,859                 & 1,015                & 36.12\%                      & 35.22\%                             & 98\%                 & 4,109                & 100.00\%                     & 21.01\%                             & 21\%                 \\
SMIE    & abusive 2            & test           & 2,442                 & 377                  & 37.84\%                      & 36.45\%                             & 96\%                 & 1,602                & 100.00\%                     & 20.64\%                             & 21\%                 \\
SMIE    & sentiment 1          & train          & 6,543                 & 999                  & 15.77\%                      & 13.48\%                             & 85\%                 & 4,269                & 100.00\%                     & 27.31\%                             & 27\%                 \\
SMIE    & sentiment 1          & test           & 1,813                 & 378                  & 15.00\%                      & 12.58\%                             & 84\%                 & 1,607                & 100.00\%                     & 27.74\%                             & 28\%                 \\
SMIE    & sentiment 2          & train          & 20,679                & 4,430                & 18.48\%                      & 16.18\%                             & 88\%                 & 30,566               & 100.00\%                     & 28.58\%                             & 29\%                 \\
SMIE    & sentiment 2          & test           & 5,719                 & 1,398                & 18.57\%                      & 16.49\%                             & 89\%                 & 8,566                & 100.00\%                     & 28.66\%                             & 29\%                 \\
SMIE    & sentiment 3          & train          & 3,601                 & 775                  & 100.00\%                     & 100.00\%                            & 100\%                & 3,829                & 100.00\%                     & 15.16\%                             & 15\%                 \\
SMIE    & sentiment 3          & test           & 1,007                 & 299                  & 99.90\%                      & 99.90\%                             & 100\%                & 1,276                & 100.00\%                     & 14.60\%                             & 15\%                 \\
SMIE    & sentiment 4          & train          & 558                   & 194                  & 98.75\%                      & 98.03\%                             & 99\%                 & 491                  & 100.00\%                     & 21.15\%                             & 21\%                 \\
SMIE    & sentiment 4          & test           & 557                   & 161                  & 99.64\%                      & 99.64\%                             & 100\%                & 522                  & 100.00\%                     & 15.26\%                             & 15\%                 \\
SMIE    & sentiment 5          & train          & 1,575                 & 27                   & 95.11\%                      & 95.05\%                             & 100\%                & 720                  & 100.00\%                     & 23.94\%                             & 24\%                 \\
SMIE    & sentiment 5          & test           & 444                   & 17                   & 97.52\%                      & 97.52\%                             & 100\%                & 317                  & 100.00\%                     & 22.75\%                             & 23\%                 \\
SMIE    & sentiment 6          & train          & 9,616                 & 2,052                & 19.21\%                      & 17.34\%                             & 90\%                 & 12,165               & 100.00\%                     & 22.56\%                             & 23\%                 \\
SMIE    & sentiment 6          & test           & 17,347                & 2,879                & 19.66\%                      & 17.89\%                             & 91\%                 & 20,456               & 100.00\%                     & 21.05\%                             & 21\%                 \\
SMIE    & uncertainity 1       & train          & 1,058                 & 389                  & 57.84\%                      & 56.71\%                             & 98\%                 & 1,390                & 100.00\%                     & 30.62\%                             & 31\%                 \\
SMIE    & uncertainity 1       & test           & 314                   & 128                  & 59.55\%                      & 58.28\%                             & 98\%                 & 402                  & 100.00\%                     & 25.80\%                             & 26\%                 \\
SMIE    & uncertainity 2       & train          & 534                   & 206                  & 44.76\%                      & 43.07\%                             & 96\%                 & 620                  & 100.00\%                     & 19.29\%                             & 19\%                 \\
SMIE    & uncertainity 2       & test           & 145                   & 65                   & 36.55\%                      & 36.55\%                             & 100\%                & 187                  & 100.00\%                     & 15.86\%                             & 16\%                 \\* \bottomrule
\caption[Downstream Data Statistics]{\textbf{Downstream Data Statistics}:  \textbf{NTUs} means unique NTUs in the dataset, \textbf{\textgreater 1 NTUs} means \% Tweets with more than 1 NTU, \textbf{\textgreater 1 $\in$ E} is \% Tweets with more than 1 NTU which exist in our Embeddings $E$, and \textbf{ $\in$ E} is \% Tweets having an NTU in $E$ only across Tweets with an NTU. SMIE = SocialMediaIE.}
\label{tab:downstream-data-stats}
\end{longtable}

\twocolumn

\section{Training Details}
\label{sec:training-details}
All models were trained on NVIDIA A100 GPUs. Our context embedding size was 200. Models were trained for maximum of 15 epochs, using eary stopping via the eval dataset. We used the \texttt{adam\_hf} optimizer in HuggingFace library \footnote{\url{https://huggingface.co/docs/transformers/main_classes/trainer\#transformers.TrainingArguments.optim}} with default learning rate of \texttt{5e-5}.

\paragraph{Downstream models} were trained with an 2 layer MLP on top of BERT or \ntulm{} embeddings. MLP hidden layer has weight matrix of size $768*768$ with a $tanh$ activation. Final layer has size $768*num\_classes$.

\paragraph{NTU embeddings} were trained on 8 NVIDIA A100 GPUs using the following config: dimension=200, learning rate=0.05, epochs=10, batch size=100,000,  batch negatives=500, uniform negatives=500, num partitions=1.